\documentclass[preprint,12pt]{elsarticle}

\usepackage{amsmath}
\usepackage{bbm}
\usepackage{bbold}
\usepackage{hyperref}

\DeclareMathOperator{\adj}{adj}
\DeclareMathOperator{\Tr}{Tr}

\begin{document}

\begin{frontmatter}

\author{Sergey Pankov\corref{cor}}
\ead{sergey.pankov@gmail.com}

\cortext[cor]{Corresponding author.}

\title{Energy-conserving intermittent-contact motion in complex models}

\address{Harik Shazeer Labs, Palo Alto, CA 94301}

\begin{abstract}

Some mechanical systems, that are modeled to have inelastic collisions, nonetheless possess energy-conserving intermittent-contact solutions, known as collisionless solutions. Such a solution, representing a persistent hopping or walking across a level ground, may be important for understanding animal locomotion or for designing efficient walking machines. So far, collisionless motion has been analytically studied in simple two degrees of freedom (DOF) systems, or in a system that decouples into 2-DOF subsystems in the harmonic approximation. In this paper we extend the consideration to a $N$-DOF system, recovering the known solutions as a special $N = 2$ case of the general formulation. We show that in the harmonic approximation the collisionless solution is determined by the spectrum of the system. We formulate a solution existence condition, which requires the presence of at least one oscillating normal mode in the most constrained phase of the motion. An application of the developed general framework is illustrated by finding a collisionless solution for a rocking motion of a biped with an armed standing torso.

\end{abstract}


\end{frontmatter}

\newcommand{\T}{\mathsf T}
\newcommand{\ul}{{\,}^0} 
\newcommand{\ult}{{\,}^0\!} 
\newcommand{\tB}{\tilde{B}}
\newcommand{\lp}{\left(}
\newcommand{\rp}{\right)}

\section{Introduction}

The possibility of a persistent motion, such as hopping or walking on a level ground, in a mechanical system with inelastic collisions may appear paradoxical, at first glance. However, no laws of mechanics are violated if all the intermittent contacts along the trajectory occur at a vanishing (at the point of contact) velocity. For a solution with purely repulsive ground reaction force, as in conventional walking or hopping, acceleration must also vanish at the contact \cite{reddy2001passive}. We call such motion collisionless \cite{gomes2005collisionless}.

Although collisionless solutions have been considered in one- \cite{chatterjee2002persistent} two- \cite{gomes2005collisionless,gomes2011walking} and three- \cite{pankov2021three} dimensional models, analytical consideration was effectively limited to 2-DOF complexity. For example, the three-dimensional bipedal walker in \cite{pankov2021three}, while nominally being a 5-DOF model, consisted only of two articulated parts, conveniently decoupling into 2-DOF subsystems in the harmonic approximation (plus a 1-DOF subsystem with a forcing term). In the analysis of those simple models it was noted that although a complete description of a harmonic system includes both normal frequencies and normal vectors, the core equations of the collisionless motion problem could be formulated exclusively in terms of the normal frequencies \cite{chatterjee2002persistent,pankov2021three}. Whether this peculiarity is a general property, not limited to low complexity models, remained unknown. 

Not every model admits a collisionless solution for any set of model parameters. However, a counting argument suggests that the dimensionality of constraints (needed to meet collisionless solution requirements) on the model parameter values is small and is independent of the model complexity \cite{pankov2021three}.

On the one hand, this may indicate the relevance of collisionless models for understanding animal locomotion \cite{alexander1989optimization}, as well as for designing efficient walking machines \cite{kashiri2018overview}. Indeed, the mechanical cost of transport (COT) -- the standard measure of locomotion efficiency equal to energy expended per weight per distance traveled -- is remarkably low for humans, dogs and other animals at walking speeds (0.08 for humans and 0.04 for dogs, according to \cite{lee2013comparative}). In contrast, the majority of man-made bipeds and quadrupeds today (with some notable special purpose built exceptions \cite{bhounsule2014low,li2016simple}) typically demonstrate an order of magnitude higher COT.

On the other hand, looking numerically for a suitable collisionless solution, in even a low-DOF non-linear model, is not a simple task \cite{gomes2011walking}. And the task may be much harder for higher-DOF models. Therefore, it is desirable to extend the analytical treatment to higher complexity models, potentially gaining useful insights along the way. 

In this work, we extend the analytical analysis to $N$-DOF linear (i.e. in harmonic approximation) models. We show that the formulation in terms of the normal frequencies is a general property of collisionless motion, not limited to low complexity models. Starting with $N = 2$ example, we conjecture a general collisionless solution existence condition. Compelling evidence in support of its correctness is then provided by analyzing a general $N$ solution in the critical region, where the collisionless solution ceases to exist. We illustrate how the search for the roots of collisionless equations can be aided by two-dimensional plots. The practical value of the harmonic approximation solution is that it can be straightforwardly evolved numerically toward a collisionless solution of a non-linear model \cite{pankov2021three}.

The paper is organized as follows. In Section \ref{termnotations} we more rigorously outline the concept of collisionless motion, problem specifications and related notations. The main equations are derived in Section \ref{derivation} for general $N$. Known solutions for $N = 2$ are reproduced in Section \ref{n=2case}. In Section \ref{n>2case} the critical region solution is analyzed. The developed approach is applied to solve a new $N = 3$ model in Section \ref{rockingn3}.  We conclude with an outlook for further research in Section \ref{discussion}. 

\section{Problem terminology and notations}
\label{termnotations}

We consider a motion of a mechanical system interacting with the ground via purely inelastic collisions and non-slipping contacts. We assume no internal collisions in the system, collisions may only occur between its parts and the ground. When a collision happens at a finite velocity, a finite amount of mechanical energy dissipates at the collision. In the rest of the paper the term collision will only be used in this context. If, on the other hand, velocity continuously vanishes at the moment of contact, so no energy gets lost, we call it an impact. Unlike a collision, an impact is time-reversible. We refer to the motion containing impacts and no collisions as collisionless.

A collisionless motion can be viewed as a temporal sequence of phases, each phase characterized by a distinct configuration of contacts and separated from the adjacent phases by impacts. Contacts constrain the system, reducing its number of DOF. The impact dimensionality is defined as the number of DOF that get constrained at the impact. In this paper we limit the consideration to a periodic collisionless motion with two phases separated by a one-dimensional impact. The phase with lower DOF is called constrained, while the other is called unconstrained. We will use a prime to distinguish quantities in the constrained phase, (e.g. $x$ in the unconstrained phase becomes $x'$ in the constrained phase). We will use $p$ superscript to represent a quantity in either of two phases, (i.e. $x^p$ can be $x$ or $x'$). If a statement is obviously applicable to both phases, we may optionally omit $p$. 

Following the literature \cite{chatterjee2002persistent,gomes2005collisionless,pankov2021three}, we restrict the collisionless solution to a particular class of periodic trajectories characterized by two symmetry points $P^p$ associated with the constrained and unconstrained phases. At $P^p$, the trajectory must be continuous and invariant under a simultaneous time reversal $T$ and a spatial transformation $S^p$ which preserves the Euclidean metric and leaves the ground invariant. By the invariance of the Lagrangian to $S^p$ and the time-reversibility of collisionless motion, the invariance to $TS^p$ extends from $P^p$ to the whole trajectory. The imposed symmetries simplify the consideration, as the number of independent variables is reduced and one only needs to consider the trajectory between $P$ and $P'$, which can then be unfolded into the full solution according to the symmetries.

Let $x$ be an $N$-dimensional coordinate vector of the unconstrained phase. Let $x_N$ be the component of $x$ that gets constrained at the impact, so $x_N$ is constant in the constrained phase. To realize a collisionless motion, in addition to $\dot x_N \to 0$ at the impact, it is also required that $\ddot x_N \to 0$, otherwise the contact cannot persist \cite{reddy2001passive}. Let $t_{imp}^p$ be the time of the impact (laying between the adjacent $P$ and $P'$) measured from the symmetry point $P^p$, and $\tau^p \equiv |t_{imp}^p|$.
In summary, the following conditions must be satisfied at the impact:
\begin{equation}
  \begin{split}
    & x(\tau) = x'(-\tau'), \\ 
    & \dot{x}(\tau) = \dot x'(-\tau'), \\
    & \ddot{x}_N(\tau) = 0. 
  \end{split}
  \label{impconds}
\end{equation}

In the rest of the section, we briefly review some matrix-related notations and conventions employed in the paper. See \ref{matnotations} for more details. 

We use columnar format for vectors. In a matrix $A$ and vector $a$: $A_i$ is the $i$th matrix row, $a_i$ is the $i$th vector element, $A^i$ is the $i$th matrix column and $A_i^j = A_{ij}$ is the matrix element at the intersection of $A_i$ and $A^j$. We use square brackets to denote a matrix, commas to separate columns and semicolons to separate rows. For example, for a $n\times m$ matrix $A$, $A = [A^1, A^2, ... , A^m] = [A_1; A_2; ... ; A_n]$ and $A^i = [A_{1i}; A_{2i}; ... ; A_{ni}]$. A matrix superscript should not be confused with a power notation, which is only used for scalar quantities, i.e. $A_{ij}^k = (A_{ij})^k$. We use $A_{(i)}$ and $A^{(j)}$ to denote matrices obtained from $A$ by removal of $A_i$ and $A^j$ respectively. Also, $A_{(ij)} \equiv A_{(i)}^{(j)}$. For a vector or matrix $a = [a_1; a_2; ... ; a_N]$, we define $\bar{a} = a_{(N)}$. We reserve $I$ for $N\times N$ identity matrix and $\mathbbm{1}$ for $N$-dimensional vector of ones.

For matrices $a$ and $b$ we denote the operations of elementwise multiplication as $a \cdot b$ (also known as Hadamard product) and elementwise division as $a/b$. The operations are defined to support broadcasting and have a higher operator precedence than ordinary matrix multiplication, e.g. $ab\cdot c = a(b\cdot c)$. We introduce these notations to reveal a peculiar and somewhat simple structure of the derived equations, that may be not obvious otherwise.

\section{Collisionless motion for linear dynamics with one-dimensional impacts}
\label{derivation}

Consider a system of $N$ linear oscillators $x = [x_1; x_2; ... ; x_N]$ with the kinetic energy $T$ and potential energy $V$ given by
\begin{equation}
  T = \frac{1}{2} \dot{x}^\T m \dot{x}, \quad
  V = \frac{1}{2} x^\T k x, 
\end{equation}
where $m$ is positive definite and $k$ is non-singular. 
The corresponding Lagrangian in the presence of an external force $F$ is 
\begin{equation}
  L = \frac{1}{2} \left( \dot{x}^\T m \dot{x} - x^\T k x \right) + x^\T F.
\end{equation}
The system is invariant under a simultaneous static shift in $x \to x + x^0$ and $F \to F + F^0$ with $F^0 = k x^0$. 
One can switch to the basis of normal coordinates $Q$ via a coordinate transformation $x = X Q$ to find
\begin{equation}
  L = \frac{1}{2c} \left( \dot{Q}^\T \dot{Q} - Q^\T \lambda\cdot Q \right) + Q^\T f,
  \label{lagrangianq}
\end{equation}
where the columns of $X$ are the eigenvectors of $m^{-1}k$ with the corresponding eigenvalues forming the spectrum vector $\lambda$, the force $f = X^\T F$, and the constant $c$ is selected such that $X_N^N=1$. We assume that all $\lambda_i$ are different (i.e. the spectrum is non-degenerate) and arranged in ascending order, for convenience. Each $\lambda_i$ corresponds to a normal mode oscillating with a frequency $\omega_i = \sqrt{\lambda_i}$. When $\lambda_i < 0$, the motion is unbounded, diverging with time as $\exp(\pm \nu_i t)$, where $\nu_i \equiv |\omega_i|$. 

As can be easily verified from the equations of motion following from the Lagrangian in Eq.(\ref{lagrangianq}), given a force $F(t) = F^\upsilon(t) = F^\upsilon(0) e^{i\upsilon t}$ oscillating with a frequency $\upsilon$, the forced oscillations $Q^\upsilon$ are
\begin{equation}
  Q^\upsilon = \frac{c f^\upsilon}{\lambda - \mathbbm{1}\upsilon^2},
\end{equation}
where $f^\upsilon = X^\T F^\upsilon$.
In the original coordinates, $x^\upsilon = X Q^\upsilon$. 
Consider now a force applied only to $x_N$, oscillating with a normal mode frequency $\omega'_i$ of the constrained system, (that is $F_{i\ne N}(t) = 0$, $F_N(t) = F^{\omega'_i}_N e^{i\omega'_it}$ and $\omega'^2_i = \lambda'_i$). Then $x^{\omega'_i}_N = 0$, while $\bar x^{\omega'_i}$ is the normal mode of the constrained system corresponding to $\lambda'_i$. Note that from Cauchy's interlace theorem \cite{fisk2005very} it follows: 
\begin{equation}
  \lambda_1 < \lambda'_1 < \lambda_2 < ... < \lambda'_{N-1} < \lambda_N .
\end{equation}
Let $\tilde x^{\omega'} = x^{\omega'}/c F_N^{\omega'}$ and $X' = [\tilde{x}^{\omega'_1}, \tilde{x}^{\omega'_2}, ... , \tilde{x}^{\omega'_{N-1}}] = X\cdot X_N (1/(\lambda\bar{\mathbbm{1}}^\T - \mathbbm{1}\lambda'^\T))$. Then the transformation between the normal coordinates of the constrained system\footnote{$X'$ does not follow the same normalization convention as $X$, so the Lagrangian of the constrained system does not have the form of Eq.(\ref{lagrangianq}), when written in $Q'$.} and the original coordinates is $x = X'Q'$, while $X'_N = 0$. If in the above analysis the constraining force also has a static component $F_N^0$ (that is $F_N(t) \to F_N(t) + F^0_N$), then the transformation becomes $x = X'Q'+x^0$, where $x^0 = (k^{-1})^NF^0_N$. We can write the expression for $X'$ and the relation $X'_N = 0$ in terms of a matrix $M$
\begin{equation}
  M_{ij} = \frac{1}{\lambda_i - \lambda'_j},
  \label{m}
\end{equation}
as 
\begin{equation}
  X' = X \cdot X_N M,
  \label{xxxnm}
\end{equation}
and 
\begin{equation}
  X_N \cdot X_N M = 0.
  \label{xnxnm}
\end{equation}
Let $\eta^\T = X_N \cdot X_N$. Since $\eta_N = X_{NN}^2 = 1$, $\eta$ can be determined from Eq.(\ref{xnxnm}):
\begin{equation}
  \bar{\eta}^\T = -M_N\bar{M}^{-1}
  = \left(\lambda' - \lambda_N \bar{\mathbbm{1}} \right)^\T \bar{M}^{-1} .
  \label{etam}
\end{equation}
Note that $M$ is a so-called Cauchy matrix, and $\bar{M}$ is a square Cauchy matrix, for which determinant and inverse are given by explicit formulas \cite{schechter1959inversion}, that are evaluated in $\mathcal{O}(N^2)$ operations. 

Normal modes are orthogonal in the sense that their evolution (with time) is governed by the dynamics of a one-dimensional harmonic oscillator. That is, the time dependence of a normal coordinate $Q_i$ can be parameterized by two constants $q_i$ and $s_i$ as $Q_i(t) = q_i g(t,\omega_i,s_i)$, where
\begin{equation}
  g(t,\omega,s) = s\cos{\omega t} + i^{\frac{s_\lambda-1}{2}} \sqrt{1-s^2}\sin{\omega t},
  \label{gdef}
\end{equation}
$s^2 \le 1$ and $s_\lambda = \textrm{sign}(\omega^2)$. 
A complete solution of the unconstrained and constrained systems, $x(t) = XQ(t)$ and $x'(t) = X'Q'(t)+x^0$ respectively, are then readily available, given $q^p$ and $s^p$ constants. In the context of our search for a collisionless solution, we will see that $s^p$ are fixed from the outset by the symmetry of the solution, while $q^p$ are determined from equations on $x^p$ and their time derivatives at the impact. Interestingly, the equations for the impact times, as well as $q^p$, can be expressed through the spectra $\lambda^p$ alone, without the use of $X^p$. This may look surprising because, for a given $q^p$, the configuration $x^p(t)$ generally depends on $X^p$. In our case, however, this dependence is confined to $\bar x^p(t)$, as follows from Eq.(\ref{etam}). In other words, $X^p$ only influences the part of the configuration that is irrelevant to the impact. 

We will use a shortened notation $g_i = g(t,\omega_i,s_i)$, so $Q = q\cdot g$. Note also $\ddot g = -\lambda\cdot g$. 
We can now write the impact conditions from Eq(\ref{impconds}) in a matrix form $Ab = 0$, where 
\begin{equation}
  A =
  \begin{bmatrix}
    X\cdot g^\T & -X'\cdot g'^\T & -(k^{-1})^N \\
    X\cdot \dot{g}^\T & -X'\cdot \dot{g}'^\T & 0 \\
    X_N\cdot \ddot{g}^\T & 0 & 0
  \end{bmatrix}
  , 
  \label{amatrix}
\end{equation}
$g = g(\tau)$, $g' = g'(-\tau')$ 
and $b = [q;q';F^0_N]$. $A$ is a $(2N+1)\times 2N$ matrix. For a nontrivial solution $b$ to exist, $A$'s rank must be at most $2N-1$. To evaluate $A$'s rank, it is convenient to bring it to a block upper triangular form by applying rank-preserving transformations. Using $m^{-1}kX=\lambda^\T \cdot X$, $c X^\T mX=I$, $\eta^\T = X_N\cdot X_N$, $\ddot g = -\lambda\cdot g$ and Eq(\ref{xxxnm}), after straightforward manipulations, one finds (see \ref{uppertriangreduction} for details):
\begin{equation}
  A \to \tilde A = 
  \begin{bmatrix}
    I & 
    \begin{bmatrix}
      M & \frac{1}{\lambda}
    \end{bmatrix} \\
    0 & B
  \end{bmatrix}
  , \quad
  B = 
  \begin{bmatrix}
    \begin{bmatrix}
       U & \frac{1}{\lambda}
    \end{bmatrix} \\
    \eta^\T \mathbbm{1} \mathbbm{1}^\T
  \end{bmatrix}
  ,
  \label{aatb}
\end{equation}
where
\begin{equation}
  U = M - G\cdot M
  , \quad 
  G =\frac{g}{\dot g} \cdot \left(\frac{\dot g'}{g'}\right)^\T
\end{equation}
The condition $\textrm{rank}(A) < 2N$ is equivalent to $\textrm{rank}(B) < N$, which under the assumption of non-singular $\bar{U}$ can be written as $\det{B_{(N)}} = \det{B_{(N+1)}} = 0$ in the form:
\begin{equation}
  \begin{bmatrix}
    \lambda_N U_N \\
    \eta^\T \mathbbm{1} \bar{\mathbbm{1}}^\T
  \end{bmatrix}
  \bar{U}^{-1}\frac{1}{\bar{\lambda}} =
  \begin{bmatrix}
    1 \\
    \eta^\T \mathbbm{1}
  \end{bmatrix}  
  .
  \label{imptimeeqs}
\end{equation}
We will call the above equations, written in this or an equivalent form, the impact equations. We would like to stress that the impact equations only involve $\tau^p$ and $\lambda^p$, thus validating our claim that the collisionless solution is determined by the spectra $\lambda^p$ alone.

As was explained in Section \ref{termnotations}, the collisionless solution is invariant to $TS^p$ associated with the symmetry point $P^p$, where $S^p$ is a Euclidean metric preserving transformation, i.e. a reflection or rotation. 
For a non-degenerate spectrum each normal mode must respect $TS$ symmetry, leaving only two possibilities at $P$: a) $q_i$ is unchanged by $S$ and $g_i(t)$ is symmetric ($|s_i|=1$), b) $q_i$ is flipped by $S$ and $g_i(t)$ is antisymmetric ($s_i=0$). Without loss of generality, we will consider non-negative $s_i$. In that case, $g_i(t)/\dot g_i(t) = \sigma_i\tan^{\sigma_i}(\omega_i t)/\omega_i$, where $\sigma_i = 1-2s_i$. Therefore,
\begin{equation}
  G_{ij} = -\frac{\sigma_i\tan^{\sigma_i}(\omega_i \tau)\omega'_j}{\sigma'_j\tan^{\sigma'_j}(\omega'_j \tau')\omega_i}.
  \label{gij}
\end{equation}

\section{$N = 2$ models}
\label{n=2case}

Eq.(\ref{imptimeeqs}) is easily solvable for $N = 2$. For $G$ one finds:
\begin{equation}
  G = \frac{\lambda'}{\lambda}.
\end{equation}
Using Eq.(\ref{gij}), we can write the impact equations as
\begin{equation}
  \frac{\sigma_i\tan^{\sigma_i}(\omega_i \tau)\omega_i}{\sigma'_1\tan^{\sigma'_1}(\omega'_1 \tau')\omega'_1} = -1, \quad i = 1, 2.
  \label{n2tteqs}
\end{equation}
Since the symmetry points are separated by an impact, we are interested in the solution with $tt' < 0$, (correspondingly, $\tau \tau' > 0$). 
Note also, for $\omega^2 < 0$
\begin{equation}
  \sigma\tan^\sigma(\omega t) \omega = -\tanh^\sigma(\nu t) \nu .
  \label{omabsom}
\end{equation}
We found that the impact equations have a non-trivial solution (i.e. a solution with finite impact times) only for
\begin{equation} 
  \lambda'_1 > 0,
  \label{n2solexist}
\end{equation}
see \ref{n2solexistence} for details. 

It is convenient to express the impact equations using dimensionless notations $o_i^p = |\omega_i^p| \tau^p$, which were called impact phases in \cite{pankov2021three}. For example, for $\lambda_i > 0$, Eq.(\ref{n2tteqs}) reads in terms of the impact phases
\begin{equation}
  \frac{\sigma_i\tan^{\sigma_i}(o_i) o_i}{\sigma'_1\tan^{\sigma'_1}(o'_1) o'_1} = -\mu , 
  \label{n2omueqs}
\end{equation}
where $\mu = \tau/\tau'$. If $\omega^2 < 0$, then $\sigma\tan^\sigma(o) o$ should be replaced with $-\tanh^\sigma(o) o$, according to Eq.(\ref{omabsom}) .

Several collisionless models with $N = 2$ have been considered in the literature. We show below how their corresponding impact equations are captured by Eq.(\ref{n2tteqs}). Their solutions can be conveniently expressed in terms of the positive roots of the equation
\begin{equation}
  \tan{y} = a \tanh{b y},
\end{equation}
defined such that $y_n(a,b) \in [(n-1)\pi,n\pi)$. We also define $\beta_n(\rho) = y_n(-\rho,\rho)$, $\gamma_n(\rho) = y_n(1/\rho,\rho)$ and $\alpha_n = \gamma_n(0^+)$. We will list the solutions in terms of the impact phases $o_2$ and $o'_1$, representing the impact times $\tau = o_2/\omega_2$ and $\tau' = o'_1/\omega'_1$.

\begin{figure}
  \centering
    \includegraphics[width=.6\columnwidth]{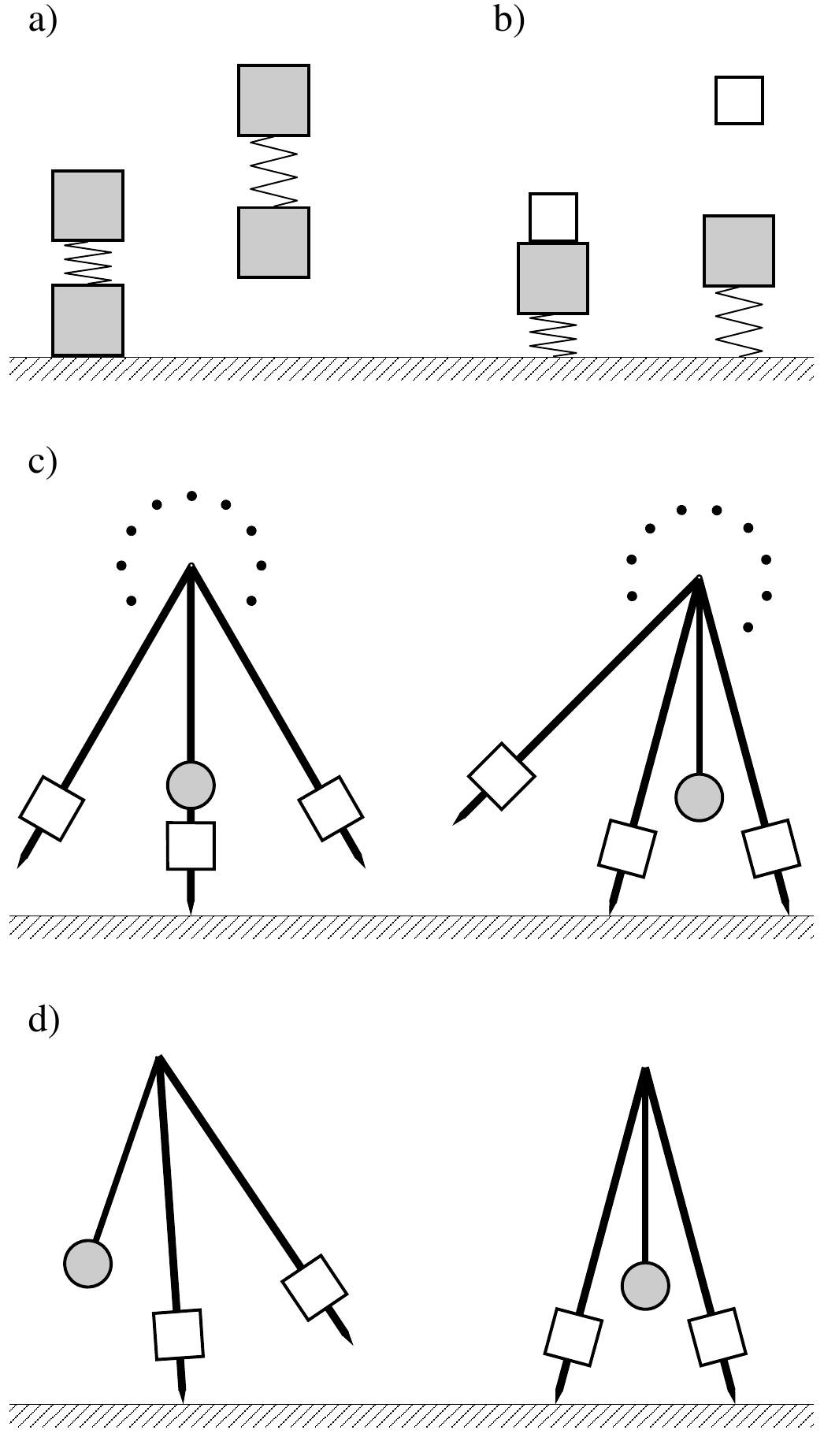}
    \caption{Schematic depiction of the models considered in Section \ref{n=2case}, with the squares and circles representing masses. a) Hopping and b) juggling. c) Extended rimless wheel. Only three spikes are displayed, with the rest shown as dots. d) Coronal bipedal rocking. In c) and d) the white squares are rigidly affixed together, forming a single part. Each model is shown in both $P^p$ configurations, with $P$ configuration shown on the right in a-b) and on the left in c-d).}
  \label{models}
\end{figure}

\subsection{Hopping and juggling}

Collisionless hopping and juggling models were studied in \cite{chatterjee2002persistent}. The hopping model consisted of two masses connected by a spring, with one of the masses undergoing impacts with the ground, see Fig.(\ref{models}.a). In the juggling model, one of the masses was connected by a spring to the ground, with the second free-falling mass undergoing impacts with the first, see Fig.(\ref{models}.b). For both models $\lambda_1 = 0$, $\sigma = [-1;-1]$ and $\sigma' = [-1]$. It follows that despite the differences in their models, the two systems have identical impact equations and identical impact times, when expressed in terms of the frequencies. For the impact equations we have:
\begin{equation}
  \begin{split}
    & o_2 \cot{o_2} = 1, \\
    & \mu o'_1 \cot{o'_1} = -1 .
  \end{split}
\end{equation}
The first equation (cf. Eq.(12) in \cite{chatterjee2002persistent}) has infinitely many roots for $o_2 > 0$, which we defined above as $\alpha_n$. For $n \to +\infty$, $\alpha_n \to (n-1/2)\pi-1/n\pi + \mathcal{O}(n^{-2})$. For a given $n$, the second equation has a single physically realizable solution. Solving for the impact phases, we find:
\begin{equation}
  \begin{split}
    & o_2 = \alpha_n, \\
    & o'_1 = \pi - \arctan{\left(\alpha_n\frac{\omega'_1}{\omega_2}\right)} .
  \end{split}
\end{equation}
For $n \to +\infty$, we have $o'_1 \to \pi/2 + (\omega_2/\omega'_1)/((n - 1/2)\pi - 1/n\pi) - (\omega_2/\omega'_1)^3/3(n\pi)^3 + \mathcal{O}(n^{-4})$.

\subsection{Extended rimless wheel}
\label{rimless}

The extended rimless wheel \cite{gomes2005collisionless} is a modification of the classical rimless wheel \cite{mcgeer1990passive} by the addition of an oscillatory DOF, in such a way that preserves the discrete rotational symmetry of the original model. In \cite{gomes2005collisionless} the additional DOF was from a reaction wheel coupled to the rimless wheel by a torsional spring. It can also be implemented by attaching a pendulum to the wheel's center, see Fig.(\ref{models}.c). We consider the model in the harmonic approximation, when the number of wheel spikes goes to infinity.

In this model $\lambda_1 < 0 < \lambda'_1$. The solution considered in \cite{gomes2005collisionless}, corresponding to a persistent rolling motion, has the following symmetry $\sigma = [1;1]$ and $\sigma' = [1]$. The impact equations are:
\begin{equation}
  \begin{split}
    & o_2 \tan{o_2} = -o_1 \tanh{o_1} , \\
    & \mu o'_1 \tan{o'_1} = o_1 \tanh{o_1} .
  \end{split}
\end{equation}
The solution is:
\begin{equation}
  \begin{split}
    & o_2 = \beta_n\left(\frac{\nu_1}{\omega_2}\right), \\
    & o'_1 = - \arctan{\left(\frac{\omega_2}{\omega'_1}\tan{\left(\beta_n\left(\frac{\nu_1}{\omega_2}\right)\right)}\right)}.
  \end{split}
\end{equation}
For $n \to +\infty$, $\beta_n(\rho) \to n\pi - \arctan{\rho} + \mathcal{O}(e^{-2\pi n\rho})$ and 
$o'_1 \to \arctan{(\nu_1/\omega'_1)} + \mathcal{O}(e^{-2\pi n\nu_1/\omega_2})$.

\subsection{Coronal rocking of a bipedal walker}
\label{rocking}

A rocking motion in the coronal plane (see Fig.(\ref{models}.d)) of a three-dimensional (kneeless) bipedal walker was used in \cite{pankov2021three} to realize a collisionless gait with finite foot-ground clearance. This is similar to the extended rimless wheel model, $\lambda_1 < 0 < \lambda'_1$, but with a different motion symmetry: $\sigma = [-1;-1]$ and $\sigma' = [1]$. The impact equations are:
\begin{equation}
  \begin{split}
    & o_2 \cot{o_2} = o_1 \coth{o_1} , \\
    & \mu o'_1 \tan{o'_1} = o_1 \coth{o_1} .
  \end{split}
\end{equation}
(Cf. the first line in Eq(55) in \cite{pankov2021three}, where $\mu$ was restricted to $1$.) 
The solution is:
\begin{equation}
  \begin{split}
    & o_2 = \gamma_n\left(\frac{\nu_1}{\omega_2}\right), \\
    & o'_1 = \arctan{\left(\frac{\omega_2}{\omega'_1} 
      \cot{\left(\gamma_n\left(\frac{\nu_1}{\omega_2}\right)\right)}\right)}.
  \end{split}
\end{equation}
For $n \to +\infty$, $\gamma_n(\rho) \to (n-1)\pi + \arctan{(1/\rho)} + \mathcal{O}(e^{-2\pi n\rho})$ and 
$o'_1 \to \arctan{(\nu_1/\omega'_1)} + \mathcal{O}(e^{-2\pi n\nu_1/\omega_2})$.

In the last two examples (the models of Sections \ref{rimless} and \ref{rocking}) we see that $\tau' \to +\infty$ for $\lambda'_{N-1} \to 0$. We show in the next section that this is a general result for $N \ge 2$, indicating that the solution existence condition of Eq.(\ref{n2solexist}) is a special case of a more general condition:
\begin{equation} 
  \lambda'_{N-1} > 0.
  \label{solexist}
\end{equation}

\section{Critical region solution for $N > 2$}
\label{n>2case}

For $N = 2$ we obtained analytical solutions (expressed via univariate functions $\beta_n$ and $\gamma_n$) of the impact equations for various symmetry point configurations $\sigma^p$, and proved the solution existence condition Eq.(\ref{solexist}). While for $N > 2$ a closed-form solution might not be available, we can still analyze the impact equations and provide compelling arguments that the same solution existence condition remains valid for all $N$. In addition, in this section, we present a number of analytical results obtained for vanishing $\lambda'_{N-1}$.

We define 
\begin{equation}
  w_i = \sigma_i \tan^{\sigma_i}(\omega_i\tau) \omega_i
  ,
  \label{wdef}
\end{equation}
and hence
\begin{equation}
  G_{ij} = -(w_i/\lambda_i)/(w'_j/\lambda'_j).
  \label{gwlam}
\end{equation}
One can observe that: a) for $\lambda'_{N-1} = 0$ and $G^{N-1} = 0$, we have $\textrm{rank}(B) < N$, b) all quantities entering $B$, apart from $w'$, are analytic in $\lambda'$, while $w'$ is analytic in $\omega'$. 
 It then follows that $w'_{N-1}$ is non-singular along the solution of the impact equations for $\lambda'_{N-1} \to 0$, i.e. 
\begin{equation}
  w'_{N-1} \to c_0 ,
  \label{wpn1c0}
\end{equation}
where $c_0$ is some constant. 
With this insight, it is straightforward to derive equations on $\tau^p$ for $\lambda'_{N-1} \to 0^+$ by differentiating $\det{B_{(N)}}$ and $\det{B_{(N+1)}}$, as we explain below. 

First, let us show that if $\lambda'_{N-1} = 0$ and $G^{N-1} = 0$, then $\textrm{rank}(B) < N$. Let $\tau^p(\lambda'_{N-1})$ represent the dependence of the impact equations' solution on $\lambda'_{N-1}$, with the rest of the spectra being fixed. For a quantity $A$ depending on $\tau^p$ and $\lambda'_{N-1}$, let $A(\lambda'_{N-1})$ be a shorthand notation for $A\left(\tau(\lambda'_{N-1}), \tau'(\lambda'_{N-1}), \lambda'_{N-1}\right)$. We define $\lambda'_{N-1} = 0$ limit as:
\begin{equation}
   \ult A = \lim_{\lambda'_{N-1} \to 0^+}
   A\left(\lambda'_{N-1}\right),
  \label{existlim}
\end{equation}
which we denote with a left superscript $0$. If we assume $\ul G^{N-1} = 0$ then $\ul U^{N-1} = \ult M^{N-1} = 1/\lambda$. Therefore, $\ult B^{N-1} = \ult B^{N}$, and hence $\textrm{rank}(\ult B) < N$. From Eqs.(\ref{wdef},\ref{wpn1c0}) it follows (for $c_0 > 0$, which remains to be shown) that $\tau' \to +\infty$ for $\lambda'_{N-1} \to 0^+$, and hence $\ul G$ and $\ul U$ are independent of $\tau'$. 
We also define a differential operator
\begin{equation}
   d_0 A = \lim_{\lambda'_{N-1} \to 0^+}
   \frac{\partial A\left(\lambda'_{N-1}\right)}{\partial \lambda'_{N-1}} .
\end{equation}
Expanding $\det{B_{(N)}} = \det{B_{(N+1)}} = 0$ around $\lambda_{N-1}' = 0$, one concludes that the necessary condition for $\textrm{rank}(B) < N$ becomes, to first order, 
\begin{equation}
  d_0\det{B_{(N)}} = d_0\det{B_{(N+1)}} = 0
  \label{d0bnbn1}
\end{equation}

Let a matrix $\tB$ have the following properties: a) $\tB$ is $N \times N$ matrix, b) $\ult \tB^{N-1} = \ult \tB^{N}$, c) $\bar{\tB} = [\bar U, 1/\bar\lambda]$. In the rest of the chapter, to simplify notations, we will be omitting the left superscript wherever the limit of Eq.(\ref{existlim}) is clearly implied. One can prove 
(see \ref{critical} for details) the following formula:
\begin{multline}
  d_0\det{\tB} = 
  -\tB_N \adj\left(\bar{U}\right)
  \frac{\bar{\mathbbm{1}}+\frac{\bar{w}}{c_0}}{\bar\lambda\cdot\bar\lambda} 
  -\det\left(\bar{U}\right)
  \left(d_0 \tB_{N,N-1} - d_0 \tB_{N,N} \right)
  .
  \label{d0detbtil}
\end{multline}
From Eq.(\ref{d0detbtil}) we find for $\tB = B_{(N)}$
\begin{equation}
  d_0\det{B_{(N)}} = 
  -\eta^\T \mathbbm{1} \bar{\mathbbm{1}}^\T \adj\left(\bar{U}\right)
  \frac{\bar{\mathbbm{1}}+\frac{\bar{w}}{c_0}}{\bar\lambda\cdot\bar\lambda}
  \label{d0detbn}
\end{equation}
and for $\tB = B_{(N+1)}$
\begin{equation}
  d_0\det{B_{(N+1)}} = 
  -U_N \adj\left(\bar{U}\right)
  \frac{\bar{\mathbbm{1}}+\frac{\bar{w}}{c_0}}{\bar\lambda\cdot\bar\lambda}
  -\det\left(\bar{U}\right)\frac{1+\frac{w_N}{c_0}}{\lambda_N^2}
  .
  \label{d0detbn1}
\end{equation}
We define
\begin{equation}
  \begin{split}
    & 
    K = 
    \begin{bmatrix}
      U_N \\
      \eta^\T \mathbbm{1} \bar{\mathbbm{1}}^\T
    \end{bmatrix}
    \frac{\adj{\bar{U}}}{\left(\bar\lambda\cdot\bar\lambda\right)^\T}
    \begin{bmatrix}
      \bar{\mathbbm{1}} & \bar{w}
    \end{bmatrix}
    , \\
    & 
    \tilde K = 
    \begin{bmatrix}
      1 \\
      0
    \end{bmatrix}
    \frac{\det{\bar{U}}}{\lambda_N^2}
    \begin{bmatrix}
    1 & w_N
    \end{bmatrix}
  .
  \end{split}
  \label{kktil}
\end{equation}
Note that $K$ and $\tilde K$ do not depend on $\tau'$, (since $\ul U$ only depends on $\tau$) . From Eqs.(\ref{d0detbn},\ref{d0detbn1},\ref{kktil}) we see that the conditions in Eq.(\ref{d0bnbn1}) can be written as $(K + \tilde K)[1;1/c_0] = 0$, from where we have the following (essentially decoupled) equations on $\tau$ and $\tau'$:
\begin{equation}
  \begin{split}
    & \det\left(K + \tilde K \right) = 0 , \\
    & c_0 = - \frac{K_{22}}{K_{21}} .
  \end{split}
  \label{critimpeqs}
\end{equation}
For $N = 2$ these equations further simplify to
\begin{equation}
  \begin{split}
    & w_1 = w_2 , \\
    & c_0 = - w_1 ,
  \end{split}
\end{equation}
which is in agreement with the results of Section \ref{n=2case}.

For a general $N$, it is also straightforward to solve for $\tau$ in the asymptotic case of large $\tau$, when the number of oscillations in the unconstrained phase is large. 
See \ref{critical} for details.

We have shown that for $\lambda'_{N-1} \to 0$, Eq.(\ref{wpn1c0}) holds for any $N \ge 2$. If $c_0 > 0$, it follows that $o'_{N-1}$ is constant in the limit $\lambda'_{N-1} \to 0^+$ (see Eq.(\ref{largetauo}) for an explicit formula) and hence $\tau' \propto 1/\omega'_{N-1}$, while no $\tau' > 0$ solution exists for $\lambda'_{N-1} \to 0^-$. For $N=2$, $c_0 = -w_1 = \tanh^{\sigma_1}(o_1)\nu_1 > 0$. For $N > 2$ we empirically verify the condition $c_0 > 0$ by randomly sampling the spectral values. While not a rigorous proof, it strongly suggests the correctness of the general collisionless solution existence condition Eq.(\ref{solexist}).

\begin{figure}
  \centering
    \includegraphics[width=.6\columnwidth]{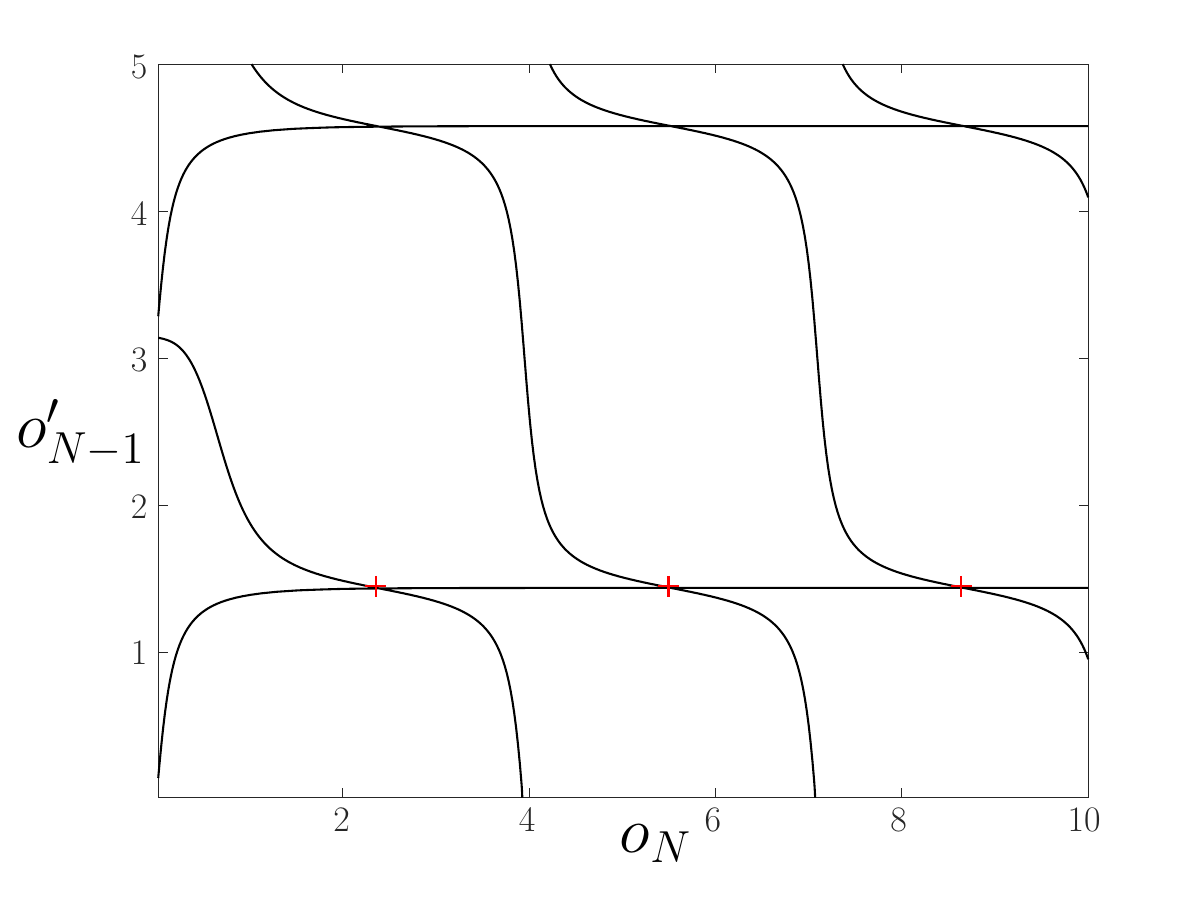}
    \caption{The zero contour of $\phi(o_N,o'_{N-1})$ from Eq.(\ref{phioo}), $N = 4$. The spectrum is random and near critical with $|\lambda_{i}| = \mathcal{O}(1)$ and $\lambda'_{N-1} = 0.01$. The curves that run (predominantly) top-to-bottom and left-to-right represent the solutions of $\det{B_{(N)}} = 0$ and $\det{B_{(N+1)}} = 0$ respectively. Their intersections are the solutions of the impact equations. The cross symbols depict the large-$\tau$ solution derived in \ref{critical}, see Eq.(\ref{largetauo}).}
  \label{impeqscont}
\end{figure}

We found it helpful to visualize the solution of the impact equations using a contour-plot in $(o_N,o'_{N-1})$ coordinates (equivalent to $(\tau,\tau')$ coordinates, up to a rescaling). We could plot a function that turns zero whenever either of the impact equations is satisfied, such as $\det{B_{(N)}}\det{B_{(N+1)}}$. However, this function is discontinuous (where $\dot g_i = 0$ or $g'_i = 0$). Instead, we will use
\begin{equation}
  \check B = B 
  \cdot [\dot g; 1]
  \cdot [g'; 1]^\T
  =
  \begin{bmatrix}
    \lp\dot g\cdot g'^\T - g\cdot\dot g'^\T \rp\cdot M 
    & \frac{\dot g}{\lambda} \\
    \eta^\T \mathbbm{1} \bar{\mathbbm{1}}^\T \cdot g'^\T & \eta^\T \mathbbm{1}
  \end{bmatrix}
  ,
  \label{checkb}
\end{equation}
which is continuous, and for which $\textrm{rank}(\check B) = \textrm{rank}(B)$ (away from the discontinuities of $B$). In Fig.(\ref{impeqscont}) we plot the zero contour of 
\begin{equation}
  \phi(o_N,o'_{N-1}) = \det{\check B_{(N)}}\det{\check B_{(N+1)}}.
  \label{phioo}
\end{equation}
Different curved lines correspond to the solutions of $\det{B_{(N)}} = 0$ and $\det{B_{(N+1)}} = 0$, while their intersections represent the solutions of the impact equations. 
Whether such a solution is physically realizable needs to be further verified by ensuring that no unwanted ground penetration happens in the unconstrained phase and no ground contact loss happens in the constrained phase. In the figure we consider a near critical configuration of a randomly sampled (and shifted) spectrum, so that $0 < \lambda'_{N-1} \ll -\lambda_{N-1}, \lambda_N$ for $N = 4$. In that regime the picture is qualitatively similar for different values of $N$, as also follows from our analysis of the critical region (in this section and \ref{critical}). We also depicted the analytical asymptotic large-$\tau$ solution presented in \ref{critical}, see Eq.(\ref{largetauo}). The asymptotic solution forms a square grid with the step size $\pi$. We only depict the bottom row of the grid, relying on the intuition from $N = 2$ case, for which one can verify that only the bottom row is physically realizable. 

\section{Collisionless rocking motion of a biped with an armed standing torso}
\label{rockingn3}

So far, we have developed a general approach for efficiently finding collisionless solutions in complex models. We have reduced the problem to a nonlinear equation in just two variables (regardless of the model complexity) -- the impact times. We also proposed a solution existence condition and showed how $N = 2$ collisionless solutions, considered in the literature, easily follow from our formulation. In this section we illustrate our approach by considering a rocking motion of a planar 3-DOF biped (i.e. $N = 3$), which to our knowledge has not been previously studied.

\begin{figure}
  \centering
    \includegraphics[width=.6\columnwidth]{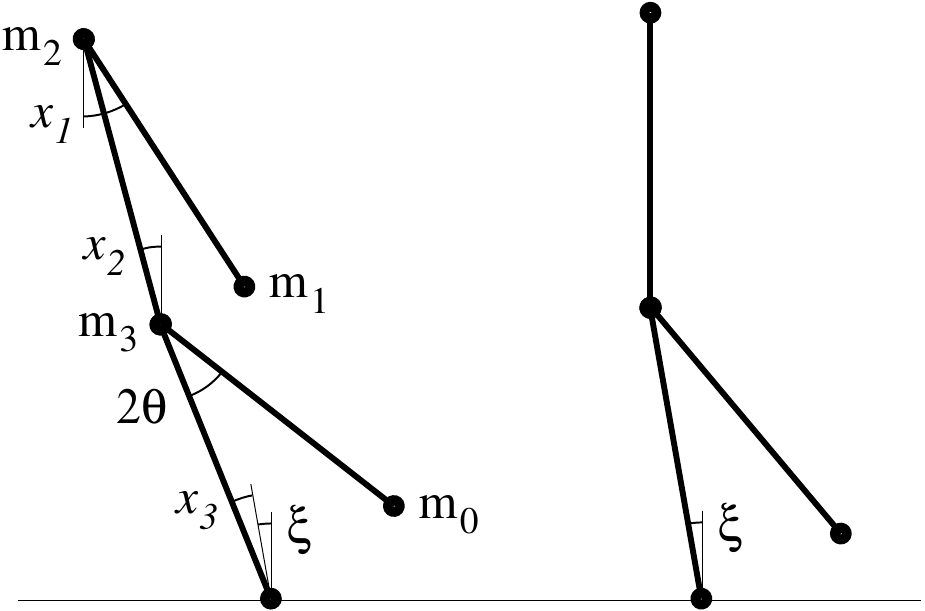}
    \caption{$N = 3$ biped with a standing armed torso. The legs are rigidly affixed together at an angle $2\theta$. Angles $x = [x_1; x_2; x_3]$ are measured relative to the static equilibrium configuration, shown on the right.}
  \label{rockmodel}
\end{figure}

The model consists of three parts (see Fig.(\ref{rockmodel})): 1) a pair of legs, rigidly affixed together at an angle $2\theta$, 2) a standing torso, attached to the legs at the hip, and 3) a hanging arm, attached at the top of the torso. For simplicity, all the links have the same length $l$. Again, we consider a harmonic approximation, which becomes exact in the limit $\theta \to 0$. 

It has been suggested in the literature to use torsion springs to prop up a standing torso in the context of collisionless walking models \cite{chatterjee2000small,gomes2011walking,pankov2021three}. In ref \cite{pankov2021three} it was also hypothesized that such springs are not strictly necessary for keeping the torso pointed up, as long as the torso is endowed with a hanging arm. An explicit collisionless solution, that we obtain below, can be viewed as a definitive proof of that hypothesis. In fact, this result holds for a torso containing any number of links stacked up on top of each other. This directly follows from our solution existence condition (without the need of an explicit solution), by noticing that a single hanging link implies $\lambda'_{N-1} > 0$. 

\newcommand{\m}{\mathrm m}

The angle conventions and masses of the model are shown Fig.(\ref{rockmodel}). For the masses we use unitalicized $\m_i$ to distinguish them from the mass matrix $m$. The total mass of the model is $\m = \m_f + \m_1 + \m_2 + \m_3$, where $\m_f = 2\m_0$ is the mass of the ``feet''. The coordinates $x = [x_1; x_2; x_3]$ are the angles of the arm, torso and the stance leg, measured relative to the static equilibrium configuration $x^{eq}$, which (when measured relative to the vertical axis) is $x^{eq} = [\pi; 0; \xi]$, where $\xi = \theta\m_f/(\m-\m_f)$. Note that $F_N^0 = \theta\m gl$ and $x^0 = [0; 0; x_N^0]$, where $-x_N^0 = \theta + \xi = \theta\m/(\m-\m_f)$.

Similar to $N = 2$ rocking motion, considered in Section \ref{rocking}, the symmetries are: $\sigma = [-1; -1; -1]$ and $\sigma' = [1; 1]$. 

The mass $m$ and spring constant $k$ matrices are:
\begin{equation}
  m = l^2
  \begin{bmatrix}
    \m_1 & -\m_1 & -\m_1\\
    -\m_1 & \m_1+\m_2 & \m_1+\m_2\\
    -\m_1 & \m_1+\m_2 & \m_1+\m_2+\m_3
  \end{bmatrix}
\end{equation}
and 
\begin{equation}
  k = gl
  \begin{bmatrix}
    \m_1 & 0 & 0\\
    0 & -\m_1-\m_2 & 0\\
    0 & 0 & -\m_1-\m_2-\m_3
  \end{bmatrix}
  .
\end{equation}
In the constrained phase $m' = m_{(NN)}$ and $k' = k_{(NN)}$.

Let us summarize the formal steps of finding a collisionless solution. First, solve for the eigensystem of $(m^p)^{-1}k^p$ to find $\lambda^p$ and $X$ (see the definition of $X$ near Eq.(\ref{lagrangianq})). Construct $M$ (see Eq.(\ref{m})) and find $X'$ (see Eq.(\ref{xxxnm})). Construct $\check B$ (see Eq.(\ref{checkb}) for $\check B$ and Eqs.(\ref{etam},\ref{gdef}) for $\eta$ and $g^p$ to that end). Plot the zero contour of $\phi(o_N,o'_{N-1})$ (see Eq.(\ref{phioo})). Use the (approximate) locations of curve intersections in the plot as an initial guess of the solution of $\det{\check B_{(N)}} = \det{\check B_{(N+1)}} = 0$ to solve that system numerically to find $\tau^p$. Given the knowledge of $\tau^p$, $g^p$ and $X^p$, construct $A$ (see Eq.(\ref{amatrix})). (Optionally, as a correctness check, verify that $\textrm{rank}(A) < 2N$). Finally, use the top left corner $(2N-1)\times(2N-1)$ submatrix of $A$ to find $q^p$ as\footnote{$q^p$ can be expressed in terms of $\tilde A$ and $x^0$ without resorting to $X^p$. Although interesting theoretically, this is irrelevant numerically.}
\begin{equation}
  \begin{bmatrix}
    q \\
    q'
  \end{bmatrix}
  =
  \begin{bmatrix}
    X\cdot g^\T & -X'\cdot g'^\T \\
    \bar{X}\cdot \dot{g}^\T & -\bar{X'}\cdot \dot{g}'^\T
  \end{bmatrix} ^{-1}
  \begin{bmatrix}
    x^0 \\
    0
  \end{bmatrix}
  .
\end{equation}
The collisionless trajectory is then given by $x = X q\cdot g$ in the unconstrained phase and $x = X' q'\cdot g' + x^0$ in the constrained phase. 

The described procedure is computationally trivial, taking a few seconds of running an Octave script \cite{octave}, with almost all of the time spent on the contour plot, needed for a reasonably good initial guess of the impact equation solution.

\begin{figure}
  \centering
    \includegraphics[width=.6\columnwidth]{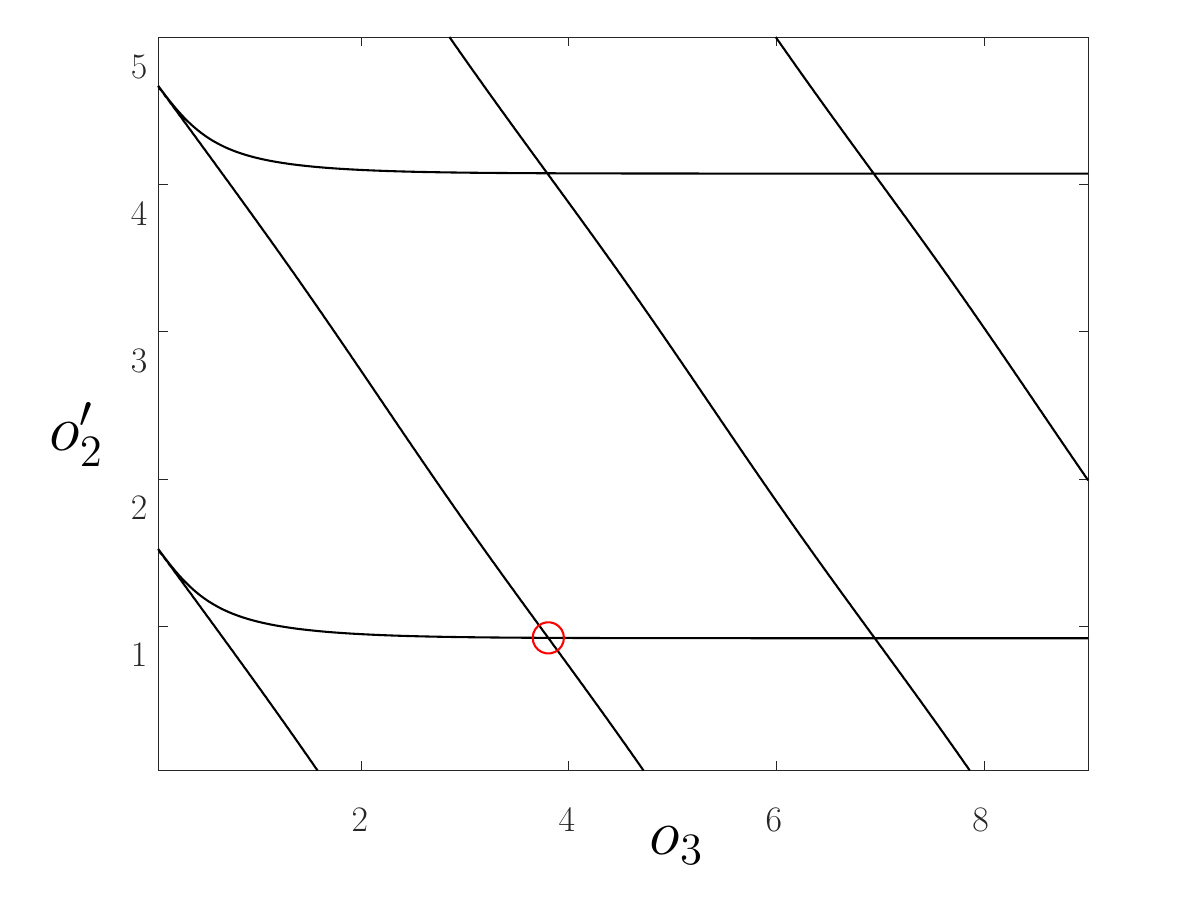}
    \caption{The zero contour of $\phi(o_N,o'_{N-1})$ for $N = 3$ rocking motion. Physically realizable collisionless solutions are represented by the bottom row of curve intersections. We focus on the solution marked by a red circle.}
  \label{rockcontour}
\end{figure}

The rest of the results in this section are presented for $g = l = \m_i|_{i=0,...,3} = 1$. Then $\xi = 2\theta/3$, $F_N^0 = 5\theta$ and $x_N^0 = -5\theta/3$. The zero-contour plot of $\phi(o_N, o'_{N-1})$ is shown in Fig.(\ref{rockcontour}). Its qualitative similarity to Fig.(\ref{impeqscont}) is not surprising, as in both cases each phase has exactly one positive eigenvalue $\lambda^p_N$, due to a hanging arm in this model. Again, the impact equation solutions appear to form a square grid asymptotically for large $\tau^p$. In the grid, columns and rows correspond to different number of oscillations in the unconstrained and constrained phases, respectively. Again, only the bottom row is physically realizable, as the impact equation solutions in the higher rows have stretches of time (in the constrained phase) with an attractive ground reaction force. We investigate a solution with the smallest number of oscillations, corresponding to the left-most intersection in the bottom row, marked with a circle in Fig.(\ref{rockcontour}). 

\begin{figure}
  \centering
    \includegraphics[width=.6\columnwidth]{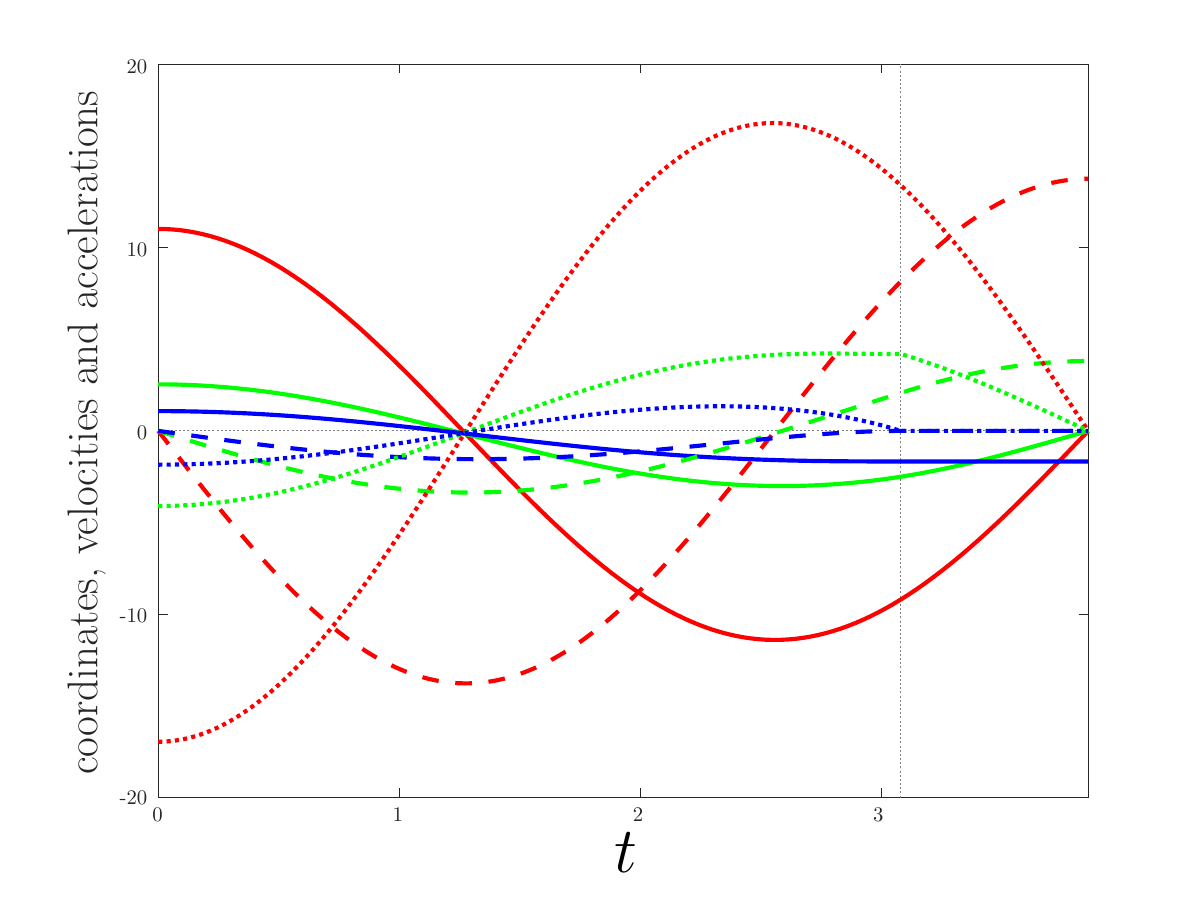}
    \caption{Complete collisionless solution, in terms of $x(t)$ (solid), $\dot{x}(t)$ (dashed) and $\ddot{x}(t)$ (dotted) plotted in the units of $\theta$, for the rocking motion of $N = 3$ armed biped, shown from $P$ (at $t = 0$) to $P'$ (at $t = \tau+\tau'$) separated by the impact (vertical gray line at $t = \tau$). Different colors are used for different components: red for $x_1$, green for $x_2$ and blue for $x_3$.}
  \label{rocksolution}
\end{figure}

\begin{figure}
  \centering
    \includegraphics[width=.6\columnwidth]{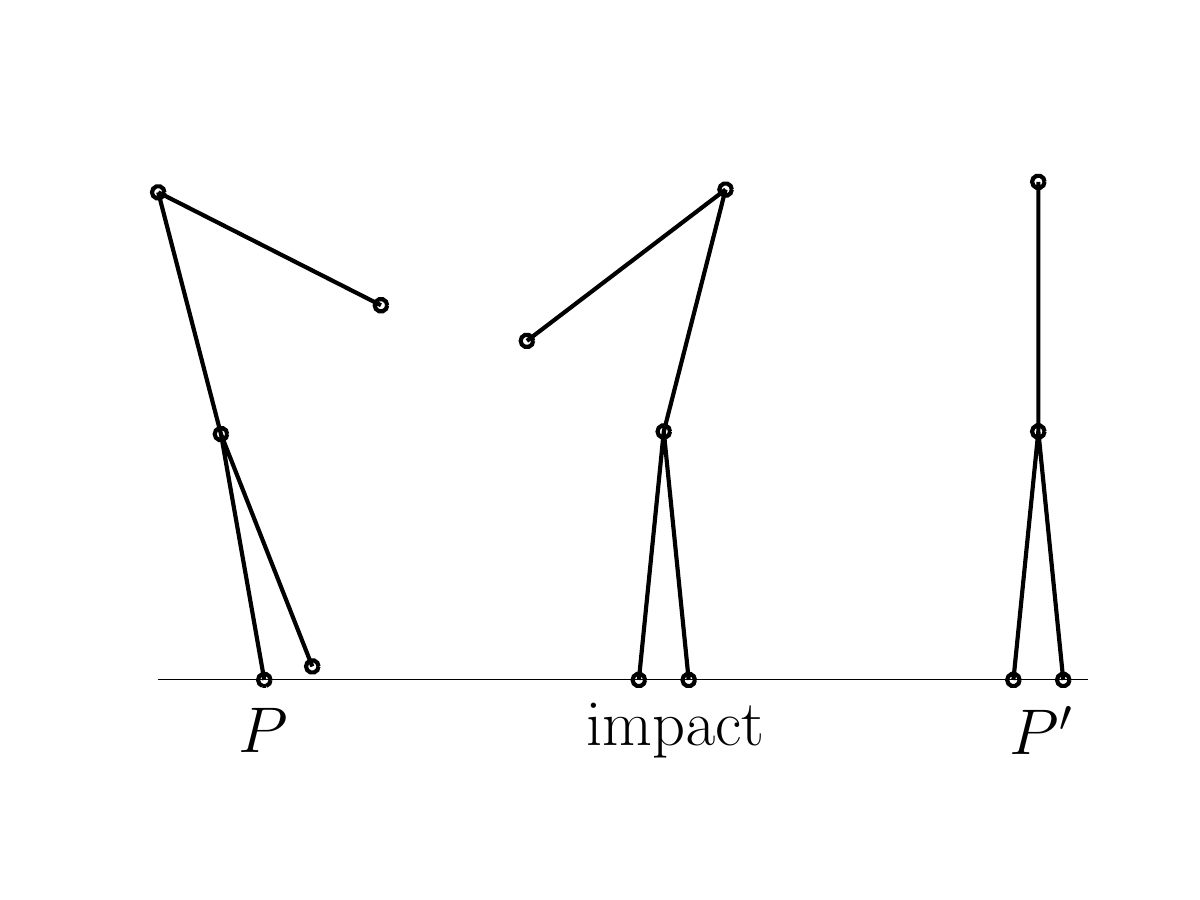}
    \caption{Visualization of the numerical solution from Section \ref{rockingn3}. The angles of the configurations are drawn to scale for $\theta = 0.1$. Left to right, the configurations are: symmetry point $P$, impact, symmetry point $P'$.}
  \label{pimpp}
\end{figure}

Solving for the impact times, as described above, we find $t_{imp} = \tau = 3.0795$ and $t'_{imp} = -\tau' = -0.77785$. See \ref{rocknumeric} for numerical values of other quantities. The complete solution $x^p(t)$ in the units of $\theta$ is shown in Fig.(\ref{rocksolution}), (also showing $\dot x^p(t)$ and $\ddot x^p(t)$). We see that, as was described in Section \ref{termnotations}, $x$, $\dot x$ and $\ddot x$ are all continuous at the impact, with $\dot x_N$ and $\ddot x_N$ vanishing, and $\ddot x$ displaying a cusp. In Fig.(\ref{pimpp}), the collisionless solution at $P$, impact, and $P'$ is depicted to scale for $\theta = 0.1$.

\section{Discussion}
\label{discussion}

In this work we have extended analytical analysis of collisionless motion to $N$-DOF models with one-dimensional impacts. The analysis revealed that the possibility of expressing the impact equations in terms of the impact phases is a general property, not limited to small $N$. The relative simplicity of this formulation enabled the investigation of the solution in the critical region, demarcated by the proposed solution existence condition.

With this development, a host of new models (e.g. multi-DOF extensions of $N = 2$ models discussed in Section \ref{n=2case}) can be further investigated. However, to be able to apply this formalism to general two-dimensional (e.g. the model of \cite{gomes2011walking}) and three-dimensional bipedal walkers, it must be further extended to cover two- and three-dimensional impacts, respectively. Our preliminary consideration suggests that in addition to relying on the constrained and unconstrained spectra, it will also involve partially-constrained spectra -- one per each dimension of the impact. 

We would like to point out how infinite friction could be used to facilitate solving for collisionless motion with higher-dimensional impacts. 
As was explained in \cite{pankov2021three}, a collisionless solution for $(n+1)$-dimensional impact requires adjustment of $n$ model parameters. For example, for a one-dimensional impact $n = 0$ (as in this paper), so a collisionless solution can be found for arbitrary model parameters, as long as the solution existence condition is met. Assume now that $n > 0$ and the acceleration vanishing (at the point of contact, e.g. foot-ground contact) is only enforced for the component orthogonal to the ground. In that case, the foot will slip right after touching the ground, and the amount of slippage will be inversely proportional to the friction coefficient $\mu_f$. Clearly, the slippage is eliminated for $\mu_f \to \infty$. Consequently, in the presence of infinite friction, there is a collisionless solution for arbitrary model parameters even for $n > 0$. In that case, collisionless solutions are easily found using the contour-plot method described in Section \ref{n>2case}. A suitable solution (e.g. with minimal number of torso oscillations) can then be continuously evolved numerically \cite{pankov2021three} back to finite $\mu_f$, with the desired topological properties of the solution preserved.

We believe that the symmetries imposed in this work are not essential for collisionless motion; they were used because of the simplifications they bring. Finding a collisionless solution without reliance on the above symmetries would be a step toward considering more faithful anthropomorphic and animal models.

While more research remains to be done, we believe that the design of complex models with desired anthropomorphic (as well as non-anthropomorphic) form factors, capable of perfectly energy-conserving mechanical motion (i.e. with zero COT), is within a reach. This paper is a contribution in that direction.

\section{Acknowledgment}

We thank Georges Harik for many useful discussions.

\appendix

\section{Matrix-related notations}
\label{matnotations}

For a matrix $A$, the matrix transposition is $A_{ij}^\T = A_{ji}$. Note that $A_i^\T = (A_i)^\T \ne (A^\T)_i$, in general. 

In Section \ref{termnotations}, we have defined the operations of elementwise multiplication $a \cdot b$ and division $a/b$ to support broadcasting: the corresponding dimensions of $a$ and $b$ must either be equal, or the smaller dimension be $1$, (the matrix element is then replicated along that dimension before the elementwise operation is applied). For example, for vectors $a$ and $b$ of dimension $n > 1$, both $a\cdot b$ and $a^\T\cdot b$ are defined, where $(a\cdot b)_i = a_i b_i$ and $a^\T\cdot b = b a^\T$. 
Note that the elementwise multiplication is commutative.

When deriving the equations involving pointwise operations, we found useful the following relations (for matrices $A$ and $B$, and a vector $c$):
\begin{equation}
  \begin{split}
    & AB\cdot c = A\cdot c^\T B \\
    & AB\cdot c^\T = (AB)\cdot c^\T
  \end{split}
\end{equation}
Other relations can be obtained from these by a transposition and relabeling of variables (e.g. $A$ to a vector $a^\T$). 

\section{Reduction of $A$ to block upper triangular form}
\label{uppertriangreduction}

Let us show how $A$ (see Eq.(\ref{amatrix})) can be reduced to a block upper triangular form $\tilde A = S_L A S_R$ given in Eq.(\ref{aatb}) by means of rank-preserving transformations $S_{L,R}$, so that $\tilde A$ depends only on $\lambda^p$ and $\tau^p$, but not on $X^p$. First of all, note that
\begin{multline}
  X^{-1} \left[ X\cdot g^\T, -X'\cdot g'^\T, -(k^{-1})^N \right] 
  = \left[ I\cdot g, -X_N^\T \cdot M \cdot g'^\T, -c X_N^\T/\lambda \right]
  .
\end{multline}
We have used Eq.(\ref{xxxnm}) to compute the second entry (in the row), and the relation $X^{-1}k^{-1} = c X^\T/\lambda$ (which follows from $m^{-1}kX=\lambda^\T \cdot X$ and $c X^\T mX=I$) to compute the third entry. We already see that the right hand side (rhs) above has no dependence on $X$ apart from $X_N$, which is related to $M$ through Eq.(\ref{etam}).

With the above observation, using rowwise and columnwise multiplications, and the Gauss eliminations, it is straightforward to bring the matrix $A$ to the form of $\tilde A$, finding along the way the expressions for $S_{L,R}$:  
\begin{equation}
  \begin{split}
  & S_L = 
  \begin{bmatrix}
    I & 0 & 0 \\
    I & I & 0 \\
    (\eta \cdot \lambda)^\T & 0 & 1
  \end{bmatrix}
  \textrm{Diag} \left( \frac{X^{-1}}{X_N^\T}, 
  -\frac{g\cdot X^{-1}}{\dot g \cdot X_N^\T}, 1 \right) , \\
  & S_R = 
  \textrm{Diag} 
  \left( \frac{X_N \cdot I}{g}, -\frac{I_{(NN)}}{g'}, -\frac{1}{c} \right) ,
  \end{split}
\end{equation}
where $\textrm{Diag}()$ is a block diagonal matrix, with the blocks listed as the arguments in parenthesis. In the derivation of $B$ in Eq.(\ref{aatb}) we have also used
\begin{equation}
  (\eta\cdot\lambda)^\T M = \eta^\T \mathbbm{1} \bar{\mathbbm{1}}^\T
  ,
\end{equation}
which is easy to verify using Eq.(\ref{xnxnm}).

\section{$N = 2$ solution existence}
\label{n2solexistence}

Below we analyze the existence of the solution of Eqs.(\ref{n2tteqs}). 
To prove that there is no solution for $\lambda'_1 < 0$ we note that the equation
\begin{equation}
  \tanh^{\sigma_1}(\nu_1 \tau)\nu_1 = -\tanh^{\sigma'_1}(\nu'_1 \tau')\nu'_1,
\end{equation}
which follows from Eq.(\ref{omabsom}) and Eq.(\ref{n2tteqs}), has no non-trivial solution (with $\tau\tau' > 0$). Consider now the equation
\begin{equation}
  \sigma_1\tan^{\sigma_1}(\omega_1 \tau)\omega_1 = \sigma_2\tan^{\sigma_2}(\omega_2 \tau)\omega_2
  \label{tteq}
\end{equation}
for $\lambda_2 > 0$. It is easy to verify that it has a solution for any $\lambda_1$. Indeed, for $\lambda_1 < 0$ we can write it as
\begin{equation}
  \sigma_1\tanh(\nu_1 \tau)\nu_1^{\sigma_1} = -\sigma_2\tan^{\sigma_1\sigma_2}(\omega_2 \tau)\omega_2^{\sigma_1}
  .
\end{equation}
The solution exists because the left hand side (lhs) is bounded and the rhs is not (bounded in either direction). Likewise, if $\lambda_1 > 0$, there exist intervals (because $\lambda_1< \lambda_2$) on which the lhs of Eq.(\ref{tteq}) is bounded, while the rhs is not. 

\section{Critical region analysis}
\label{critical}

The derivations in this section rely on the properties of $\tilde B$ listed in Section \ref{n>2case}. 
Let us first derive Eq.(\ref{d0detbtil}). 
Using that $\adj(\tB)_{ij}|_{i<N-1} = d_0\tB_{iN}|_{i<N} = 0$, we can write:
\begin{multline}
  d_0 \det{\tB} = \Tr\lp\adj\lp\tB\rp d_0\tB\rp \\ 
  = \sum_{i<N}\adj\lp\tB\rp_{N-1,i} d_0\tB_{i,N-1} 
  + \sum_{k=0,1}\adj\lp\tB\rp_{N-k,N} d_0\tB_{N,N-k}
  .
  \label{d0detbt}
\end{multline}
For the first term in the rhs of Eq.(\ref{d0detbt}) we find
\begin{multline}
  \sum_{i<N}\adj\lp\tB\rp_{N-1,i} d_0\tB_{i,N-1} 
  = \sum_{i<N} (-1)^{N-1+i} \det\lp\tB_{(iN)}\rp d_0\tB_{i,N-1} \\
  = \sum_{i,j<N} (-1)^{i+j-1} \tB_{Nj} \det\lp\bar{U}_{(ij)}\rp d_0\tB_{i,N-1} 
  = -\sum_{i,j<N} \tB_{Nj} \adj\lp\bar{U}\rp_{ji} d_0\tB_{i,N-1} \\
  = -\tB_{N} \adj\lp\bar{U}\rp d_0\bar{\tB}^{N-1}
  .
  \label{d0detterm1}
\end{multline}
For the second term in the rhs of Eq.(\ref{d0detbt}) we find
\begin{multline}
  \sum_{k=0,1}\adj\lp\tB\rp_{N-k,N} d_0\tB_{N,N-k} 
  = \det\lp\bar{U}\rp \lp d_0\tB_{N,N} - d_0\tB_{N,N-1} \rp
  .
  \label{d0detterm2}
\end{multline}
Note that for $\tB = B_{(N+1)}$ 
\begin{multline}
  d_0\tB^{N-1} = d_0 U^{N-1} 
  = d_0 M^{N-1} - d_0 G^{N-1} \cdot M^{N-1}
  = \frac{\mathbbm{1} + \frac{w}{c_0}}{\lambda\cdot\lambda}
  .
  \label{d0btn1}
\end{multline}
Combining Eqs.(\ref{d0detbt}-\ref{d0btn1}), we arrive at Eq.(\ref{d0detbtil}).
For $\tB = B_{(N)}$ we have
\begin{equation}
  d_0\tB_{N,N} - d_0\tB_{N,N-1} = 0
  \label{d0bb}
\end{equation}
Then Eq.(\ref{d0detbn}) and Eq.(\ref{d0detbn1}) readily follow from Eqs.(\ref{d0detbtil},\ref{d0btn1},\ref{d0bb}).

The critical impact equations for $\tau \to +\infty$ can be solved explicitly, because in that case the time dependence enters solely via $w_N$, which in turn only enters in $G_N$. Specifically, for $\tau^p \to +\infty$, for any $\lambda^p_i < 0$, $w^p_i \to -\nu^p_i$. Then (see Eq.(\ref{gwlam})) we can write $G_{ij} = -\nu'_j/\nu_i = -\sqrt{\lambda'_j/\lambda_i}$ and hence
\begin{equation}
  \bar U_{ij} = \frac{1+\sqrt{\frac{\lambda'_j}{\lambda_i}}}{\lambda_i-\lambda'_j} .
\end{equation}
Note that $G_N = w_N \nu'^\T/\lambda_N$ and $U_N = M_N + w_N M_N\cdot\nu'^\T/\lambda_N$. Similarly to Eq.(\ref{kktil}) we define
\begin{equation}
  R = 
  \begin{bmatrix}
    M_N\cdot\frac{\nu'^\T}{\lambda_N} \\
    M_N \\
    \eta^\T \mathbbm{1} \bar{\mathbbm{1}}^\T
  \end{bmatrix}
  \frac{\adj{\bar{U}}}{\left(\bar\lambda\cdot\bar\lambda\right)^\T}
  \begin{bmatrix}
    \bar{\mathbbm{1}} & -\bar{\nu}
  \end{bmatrix}
  , \quad
  r = \frac{\det{\bar{U}}}{\lambda_N^2}
  .
  \label{rr}
\end{equation}
Let $e$ be a $2\times 2$ identity matrix. Then the solution of (the first equation in) Eq.(\ref{critimpeqs}) can be written using the definitions of Eq.(\ref{rr}) as
\begin{equation}
  w_N = -\frac{\det\lp R_{(1)} + r e_1^\T {e_1}\rp}
  {\det\lp R_{(2)} + r e_1^\T {e_2}\rp}
  ,
  \label{wninftau}
\end{equation}
while the second line in Eq.(\ref{critimpeqs}) becomes
\begin{equation}
  c_0 = - \frac{R_{32}}{R_{31}} .
  \label{c0inftau}
\end{equation}
The expressions for $o_N$ and $o'_{N-1}$ (and thus for $\tau$ and $\tau'$) then follow from Eqs.(\ref{wdef},\ref{wninftau},\ref{c0inftau}):
\begin{equation}
  \begin{split}
    & o_N = \lp n-\frac{1-\sigma_N}{4} \rp\pi + \arctan{\frac{w_N}{\omega_N}} , \\
    & o'_{N-1} = \frac{3-\sigma'_{N-1}}{4} \pi - \frac{\omega'_{N-1}}{c_0} ,
  \end{split}
  \label{largetauo}
\end{equation}
where $n \ge 1$.

\section{Details of numerical solution for $N = 3$ collisionless rocking motion}
\label{rocknumeric}

We list below the numerical values of quantities that are obtained in the process of computation of a collisionless solution for the problem (and model parameters) in Section \ref{rockingn3}, but which were omitted in the section for brevity.

The unconstrained and constrained spectra are
\begin{equation}
  \lambda = 
  \begin{bmatrix}
    -5.85028 \\
    -0.67319 \\
    1.52348
  \end{bmatrix}
  , \quad
  \lambda' = 
  \begin{bmatrix}
    -1.4142 \\
    1.4142
  \end{bmatrix}
  ,
\end{equation}
from where $M$ directly follows
\begin{equation}
  M = 
  \begin{bmatrix}
    -0.22542 & -0.13766 \\
    1.34949 & -0.47906 \\
    0.34040 &  9.15225
  \end{bmatrix}
  .
\end{equation}
The normal modes are (appropriately normalized, so that $c = 0.019816$ in Eq.(\ref{lagrangianq})):
\begin{equation}
  X = 
  \begin{bmatrix}
    -2.3698 &  2.2804 &  9.4927 \\
    -9.3017 &  3.0480 &  2.2617 \\
    6.5268 &  2.6199 &  1
  \end{bmatrix}
  .
\end{equation}
Correspondingly, for the constrained phase:
\begin{equation}
  X' = 
  \begin{bmatrix}
    14.780 &  86.146 \\
    25.232 &  25.232 \\
    0 & 0
  \end{bmatrix}
  .
\end{equation}
Finally, the normal mode weights (in the units of $\theta$) are:
\begin{equation}
  q = 
  \begin{bmatrix}
    -0.000031265 \\
    -0.034423 \\
    1.1687
  \end{bmatrix}
  , \quad
  q' = 
  \begin{bmatrix}
    -0.0087462 \\
    0.1357027
  \end{bmatrix}
  .
\end{equation}


\bibliography{paper_arxiv.bbl}

\begin{thebibliography}{10}
\expandafter\ifx\csname url\endcsname\relax
  \def\url#1{\texttt{#1}}\fi
\expandafter\ifx\csname urlprefix\endcsname\relax\def\urlprefix{URL }\fi
\expandafter\ifx\csname href\endcsname\relax
  \def\href#1#2{#2} \def\path#1{#1}\fi

\bibitem{reddy2001passive}
C.~Reddy, R.~Pratap, A passive hopper with lossless collisions, in: IUTAM
  Symposium on Nonlinearity and Stochastic Structural Dynamics, Springer, 2001,
  pp. 209--220.

\bibitem{gomes2005collisionless}
M.~W. Gomes, Collisionless rigid body locomotion models and physically based
  homotopy methods for finding periodic motions in high degree of freedom
  models, Cornell University, 2005.

\bibitem{chatterjee2002persistent}
A.~Chatterjee, R.~Pratap, C.~Reddy, A.~Ruina, Persistent passive hopping and
  juggling is possible even with plastic collisions, The International Journal
  of Robotics Research 21~(7) (2002) 621--634.

\bibitem{gomes2011walking}
M.~Gomes, A.~Ruina, Walking model with no energy cost, Physical Review E 83~(3)
  (2011) 032901.

\bibitem{pankov2021three}
S.~Pankov, Three-dimensional bipedal model with zero-energy-cost walking,
  Physical Review E 103~(4) (2021) 043003.

\bibitem{alexander1989optimization}
R.~Alexander, Optimization and gaits in the locomotion of vertebrates,
  Physiological reviews 69~(4) (1989) 1199--1227.

\bibitem{kashiri2018overview}
N.~Kashiri, A.~Abate, S.~J. Abram, A.~Albu-Schaffer, P.~J. Clary, M.~Daley,
  S.~Faraji, R.~Furnemont, M.~Garabini, H.~Geyer, et~al., An overview on
  principles for energy efficient robot locomotion, Frontiers in Robotics and
  AI 5 (2018) 129.

\bibitem{lee2013comparative}
D.~V. Lee, T.~N. Comanescu, M.~T. Butcher, J.~E. Bertram, A comparative
  collision-based analysis of human gait, Proceedings of the Royal Society B:
  Biological Sciences 280~(1771) (2013) 20131779.

\bibitem{bhounsule2014low}
P.~A. Bhounsule, J.~Cortell, A.~Grewal, B.~Hendriksen, J.~D. Karssen, C.~Paul,
  A.~Ruina, Low-bandwidth reflex-based control for lower power walking: 65 km
  on a single battery charge, The International Journal of Robotics Research
  33~(10) (2014) 1305--1321.

\bibitem{li2016simple}
Q.~Li, G.~Liu, J.~Tang, J.~Zhang, A simple 2d straight-leg passive dynamic
  walking model without foot-scuffing problem, in: 2016 IEEE/RSJ International
  Conference on Intelligent Robots and Systems (IROS), IEEE, 2016, pp.
  5155--5161.

\bibitem{fisk2005very}
S.~Fisk, A very short proof of cauchy's interlace theorem for eigenvalues of
  hermitian matrices, American Mathematical Monthly 112~(2) (2005) 118--118.

\bibitem{schechter1959inversion}
S.~Schechter, On the inversion of certain matrices, Mathematical Tables and
  Other Aids to Computation 13~(66) (1959) 73--77.

\bibitem{mcgeer1990passive}
T.~McGeer, et~al., Passive dynamic walking, I. J. Robotic Res. 9~(2) (1990)
  62--82.

\bibitem{chatterjee2000small}
A.~Chatterjee, M.~Garcia, Small slope implies low speed for mcgeer's passive
  walking machines, Dynamics and Stability of Systems 15~(2) (2000) 139--157.

\bibitem{octave}
J.~W. Eaton, D.~Bateman, S.~Hauberg, R.~Wehbring,
  \href{https://www.gnu.org/software/octave/doc/v4.2.1/}{{GNU Octave} version
  4.2.1 manual: a high-level interactive language for numerical computations}
  (2017).
\newline\urlprefix\url{https://www.gnu.org/software/octave/doc/v4.2.1/}

\end{thebibliography}

\end{document}